\documentclass{article}

\usepackage{arxiv}

\usepackage[utf8]{inputenc} 
\usepackage[T1]{fontenc}    
\usepackage{hyperref}       
\usepackage{url}            
\usepackage{booktabs}       
\usepackage{amsfonts}       
\usepackage{nicefrac}       
\usepackage{microtype}      
\usepackage{lipsum}
\usepackage{graphicx}

\usepackage{times}
\usepackage{helvet}
\usepackage{courier}
\usepackage{color}
\usepackage{amsmath}
\usepackage{diagbox}
\usepackage{multirow}
\usepackage{booktabs}

\title{A Learning-based Discretionary Lane-Change Decision-Making Model with Driving Style Awareness}

\author{
 Yifan Zhang \\
  Department of Computing Science\\
  City University of Hong Kong\\
  \texttt{yif.zhang@my.cityu.edu.hk} \\
   \And
 Qian Xu \\
  Department of Information Engineering\\
  Zhejiang University of Technology\\
  \texttt{qianxu@zjut.edu.cn} \\
  \And
 Jianping Wang \\
  Department of Computing Science\\
  City University of Hong Kong\\
  \texttt{jianwang@cityu.edu.hk} \\
  \And
 Kui Wu \\
  School of Coumputing and Information\\
  University of Victoria\\
  \texttt{wkui@uvic.ca} \\
  \And
 Zuduo Zheng \\
  School of Coumputing and Information\\
  The University of Queensland\\
  \texttt{zuduo.zheng@uq.edu.au} \\
  \And
 Kejie Lu \\
  School of Coumputing and Information\\
  University of Puerto Rico\\
  \texttt{kejie.lu@upr.edu} \\
}

\begin{document}
\maketitle
\begin{abstract}
Discretionary lane change (DLC) is a basic but complex maneuver in driving, which aims at reaching a faster speed or better driving conditions, e.g., further line of sight or better ride quality. Although many DLC decision-making models have been studied in traffic engineering and autonomous driving, the impact of human factors, which is an integral part of current and future traffic flow, is largely ignored in the existing literature. In autonomous driving, the ignorance of human factors of surrounding vehicles will lead to poor interaction between the ego vehicle and the surrounding vehicles, thus, a high risk of accidents. The human factors are also a crucial part to simulate a human-like traffic flow in the traffic engineering area. In this paper, we integrate the human factors that are represented by driving styles to design a new DLC decision-making model. Specifically, our proposed model takes not only the contextual traffic information but also the driving styles of surrounding vehicles into consideration and makes lane-change/keep decisions. Moreover, the model can imitate human drivers' decision-making maneuvers to the greatest extent by learning the driving style of the ego vehicle. Our evaluation results show that the proposed model almost follows the human decision-making maneuvers, which can achieve $98.66\%$ prediction accuracy with respect to human drivers' decisions against the ground truth. Besides, the lane-change impact analysis results demonstrate that our model even performs better than human drivers in terms of improving the safety and speed of traffic. 
\end{abstract}


\section{Introduction}
Lane change is an essential activity in driving and can be either mandatory or discretionary. Mandatory lane change (MLC) is necessary for many scenarios, e.g., lane merging and lane drop. On the other hand, discretionary lane change (DLC) is primarily motivated by gaining a better driving condition, e.g., higher speed or better line of sight. Compared with MLC, DLC is more challenging because it can happen at any location and it is more complex with a higher degree of uncertainty~\cite{zheng2014recent}. According to the report from U.S. national highway traffic safety administration~\cite{Hs2009AnalysisOL}, about $240,000$ to $610,000$ lane-change crashes occurred annually, most of which are due to improper DLC. 

In general, DLC can be divided into three stages: (1) decision-making, (2) trajectory planning, and (3) path tracking~\cite{8648365}. Among these stages, the first one is critical to the following stages because the decision-making model determines whether to change lane and, if the change is necessary, determines when to start changing lane. In this paper, we will focus on the DLC decision-making model.

In recent years, numerous DLC decision-making models have been studied and developed in traffic engineering~\cite{zheng2014recent,ALI2020105463,singh2011estimation} and autonomous driving~\cite{8648365,8950329,8832120}. These DLC decision-making models can be classified into model-based models and learning-based models. 

Model-based DLC decision-making models apply a pre-defined mathematical formula, such as gap acceptance, to compute the lane-change probability under the current traffic conditions. Typical model-based DLC decision-making models include Gipps-type models~\cite{gipps1986model,kesting2007general}, utility theory-based models~\cite{ahmed1996models}, cellular automata-based models~\cite{maerivoet2005cellular}, game theory-based models~\cite{ALI2020105463,GameTheory2}, and Markov process-based models~\cite{singh2011estimation}. The key issues of this type of methods include:
\begin{itemize}
    \item These lane-change models in the traffic engineering literature are developed for understanding human drivers' lane-change decision mechanisms and for estimating the aggregated impact of lane change on traffic flow, and as such, they cannot be directly adopted in autonomous driving due to their mediocre prediction accuracy.
    \item Many models, except the game theory-based models, treat the lane-change decision process as a one-player decision-making event, and thus are unable to capture the interactions between the lane-changer and surrounding vehicles. 
    \item The rules and criteria defined in most existing models cannot evolve according to the changing environment, which may lead to unstable and even dangerous interactions with adjacent vehicles.
    \item The decision-making moment is often wrongly labeled as the crossing line time while the decision has been made before that. If a vehicle follows such models for decision-making, its space to the preceding vehicle may keep decreasing from the decision-making moment to the crossing line moment, which may lead to collisions.
\end{itemize}

With the development of deep learning and the emergence of the human drivers' trajectory dataset, such as the NGSIM~\cite{NGSIM} and highD datasets~\cite{highDdataset}, lots of learning-based models~\cite{8648365,8950329,8500556,XIE201941} have been proposed to learn DLC from human drivers, which attempt to overcome the limitations of model-based approaches. Moreover, a human-like DLC decision-making model will have better interactions with surrounding human-driven vehicles. Nevertheless, the existing human-like DLC decision-making models have the following limitations:
\begin{itemize}
    \item None of the current work considers the human factors of surrounding vehicles. In the mixed traffic flow containing both human-driven vehicles and autonomous vehicles, which will be the norm in the near future, human factors of surrounding vehicles, such as drivers' experiences and personality that can be expressed by driving styles, play a major role in making a proper DLC decision. Though some studies~\cite{8500333,HUO202062} consider the human factors of the ego vehicle in designing and implementing advanced driver assistant systems (ADAS), their main purpose is to detect the intention of human drivers and warn them of potential risks. 
    \item None of the existing learning-based models distinguishes DLC from MLC, despite their distinctively different nature. It can be expected that using the same learning model for both MLC and DLC may lead to lower prediction accuracy and improper lane-change decision. Moreover, many existing studies~\cite{8648365,8950329,binary_lc} assume that a target lane is given, which is often not true in practice because a driver can have multiple target lane options on a road with more than two lanes. Clearly, such an assumption may lead to unbalanced use of all available lanes and eventually may decrease traffic efficiency.
    
\end{itemize}

To address the aforementioned issues, we design and implement a learning-based driving style aware DLC (DSA-DLC) decision-making model, which is capable of (1) imitating the human driver's lane-change maneuver, (2) making the lane-change decision at a proper moment without fixing the target lane, and (3) improving the safety and speed of traffic. Specifically, we take human factors into account by adding the driving styles of surrounding vehicles as well as the ego vehicle into our DSA-DLC decision-making model. Moreover, the traffic factors, such as the space headway to the preceding vehicle, which are used in most of the current works to analyze the lane-change decision, are extended and leveraged to construct the model. Our DSA-DLC decision-making model makes full use of the above information to gauge the potential risks of lane change and then mitigate them for safety assurance like a human. Thus, under the same traffic conditions, the DSA-DLC decision-making model is capable of making almost the same lane-change decision at the same location and the same moment as human drivers. Besides its obvious application in autonomous driving, this model can also be used as the lane-change model to simulate a human-like traffic flow.

Our main contributions are summarized as follows:
\begin{itemize}
    \item We propose a novel DSA-DLC decision-making model to predict the DLC decision that is made by human drivers, which can be applied in both the autonomous driving system and traffic flow simulators.
    \item We incorporate driving styles of surrounding vehicles and accordingly select a suitable driving style for the ego vehicle through the observed trajectories.
    \item We extend the lane-change decisions into left lane change, right lane change, and lane keep, which is more general with a higher practical value and can increase the traffic efficiency in real traffic scenarios.
    \item We evaluate the DSA-DLC decision-making model by comparing the accuracy among different models and analyzing the lane-change impact on the following vehicle in the target lane from the perspectives of speed and safety.
\end{itemize}

\begin{figure}[t]
    \centering
    \includegraphics[width=13cm]{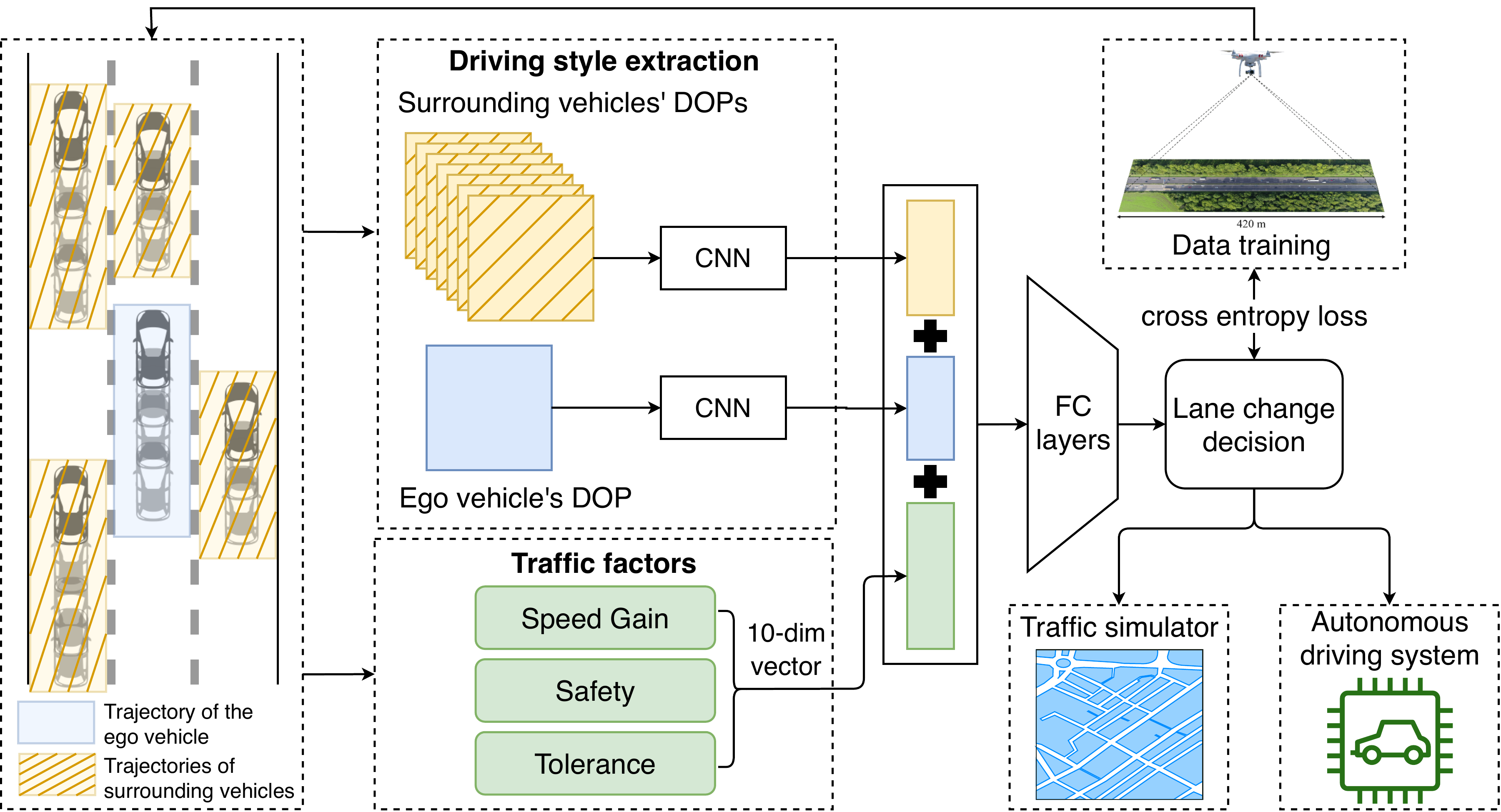}
    \caption{The architecture of DSA-DLC decision-making model.}
    \label{fig:lc_model}
\end{figure}

\begin{figure}[t]
    \centering
    \includegraphics[width=10cm]{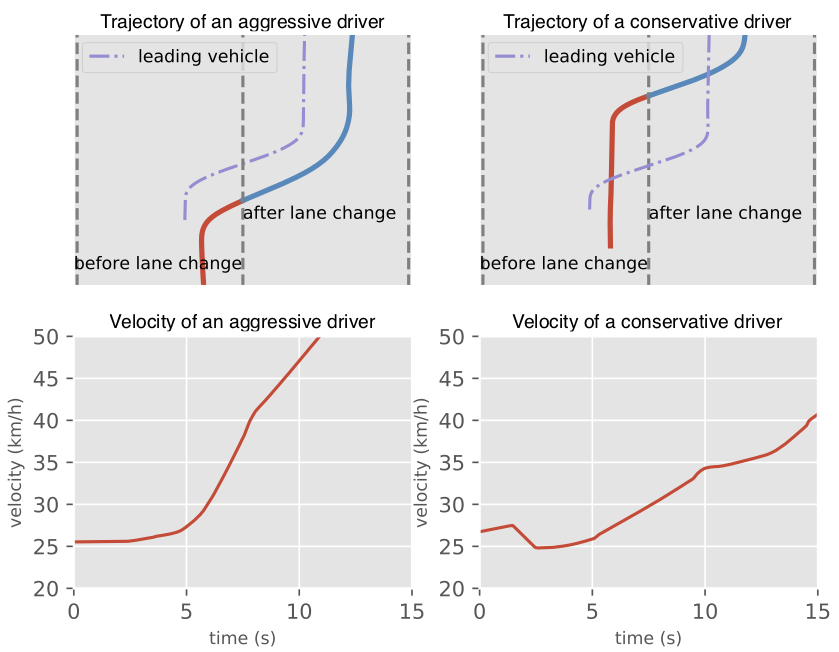}
    \caption{An example to illustrate the difference between an aggressive driver and a conservative driver in lane-change behavior.} 
    \label{fig:drivingstyle}
\end{figure}

\section{A Novel DLC Decision-Making Model with Driving Style Awareness} \label{sec:model}

To learn human drivers' DLC decision-making strategies and to utilize them to mitigate the potential risks of surrounding vehicles, we design a novel DSA-DLC decision-making model, which considers not only the contextual traffic information, but also the driving styles of the ego vehicle and surrounding vehicles. Specifically, we consider the driving styles of surrounding vehicles to reduce the potential safety risk, and use the driving style of the ego vehicle to imitate the human-driver. The main architecture of the DSA-DLC decision-making model is illustrated in Figure~\ref{fig:lc_model}.

In the rest of this section, we first analyze the driving style of a vehicle and represent it in a proper format for further utilization. Then, we explore the traffic factors that affect the lane-change decisions. Finally, we present our DSA-DLC decision-making model that outputs the lane-change/keep decisions given the aforementioned input.

\subsection{Driving Style Analysis and Representation}

Following \cite{sharma2017human} that shows how human factors can influence the lane-change decisions, we first demonstrate the impact of human factors on lane-change decisions using an example in Figure~\ref{fig:drivingstyle}, in which two drivers face the same lane-change scenario. This example is extracted from the CAT driving dataset~\cite{SHARMA2020100127}, which contains $78$ volunteers' driving trajectories in the same simulated driving environment. The information of each driver, such as age, gender, education, and driving experience, is collected by questionnaires and summarized as an aggressiveness index. Based on the aggressiveness indexes, we select an aggressive driver and a conservative driver. Figure~\ref{fig:drivingstyle} shows that, under the same surrounding environment, the aggressive driver makes a lane-change decision at the moment when the preceding vehicle is taking lane-change actions, while the conservative driver maintains a safe distance and keeps driving in the current lane until the preceding vehicle has completed the lane change. From Figure~\ref{fig:drivingstyle}, we can see that human factors' impact on lane changes can be reflected by different trajectories and speed distributions. Since human driving styles may not be truthfully obtained through questionnaires in practice, we will use the observed trajectories of the surrounding vehicles and the ego vehicle to extract driving styles corresponding to the human factors. 

Instead of classifying the driving styles into explicit categories such as aggressive and conservative, we implicitly describe the driving styles through extracting some specific features from the historical trajectory. Specifically, we apply the driving operational picture (DOP)~\cite{DOP} to describe the features that affect the driving style of a vehicle. Since the driving style may vary under different traffic conditions, we study the short-term driving style of vehicles according to the observation of historical trajectories. The time window is set to $2$ seconds, a typical response time widely used in the traffic flow modeling literature~\cite{ALI2020105463,Sharma2019EstimatingAC}. Denote matrix $D^k$ as the DOP of the vehicle $k$. $D^k$ consists of $7$ statistic features $S$ (in columns) and $8$ vehicles features $F$ (in rows). The statistic features $S = \{s_1,\ldots, s_7\}$ correspond to mean, standard deviation, median, $25\%$ percentile, $75\%$ percentile, minimum and maximum, respectively. The $8$ vehicle features $F$ are listed in Table~\ref{tab:DOP}. Hence, $D^k$ can be represented as an $8\times 7$ matrix $[e_{ij}]_{8\times 7}$. For instance, the element $e_{11}$ is the mean of the relative local $y$ (the first feature listed in Table~\ref{tab:DOP}) in the past $2$-second trajectory.

Our lane-change model involves the ego vehicle $E$, and up to $7$ closest surrounding vehicles, denoted as (\textit{P},  \textit{PL}, \textit{PR}, \textit{FL}, \textit{FR}, \textit{ASL}, \textit{ASR}), including the preceding vehicle in the current lane \textit{P}, the preceding vehicle in the left adjacent lane \textit{PL}, the preceding vehicle in the right adjacent lane \textit{PR}, the following vehicle in the left adjacent lane \textit{FL}, the following vehicle in the right adjacent lane \textit{FR}, the alongside vehicle in the left adjacent lane \textit{ASL}, and the alongside vehicle in the right adjacent lane \textit{ASR}. Note that \textit{ASL} and \textit{ASR} vehicles are those vehicles adjacent to the ego vehicle, and for safety reasons, the ego vehicle should not make any lane-change decision if there are \textit{ASL} and \textit{ASR} vehicles. The DOPs of the involved vehicles denoted as $D$ is composed of eight $D^{k}_{8\times7}$ matrices. If there is no such a surrounding vehicle, its DOP will be set as a zero matrix with a dimension of $8\times7$.

\begin{table*}
    \centering
    \begin{tabular}{p{110pt}p{60pt}p{260pt}}
    \toprule
    \textbf{Feature} & \textbf{Units} & \textbf{Description} \\
    \midrule
    Relative Local $y$ & meters & Lateral ($y$) coordinate of the front-center of the vehicle in meters with respect to the start point of the trajectory. \\
    Relative Local $x$ & meters & Longitudinal ($x$) coordinate of the front-center of the vehicle with respect to the start point of the trajectory in the direction of travel.\\
    Lateral Velocity & meters/second & Lateral instantaneous velocity of vehicle. \\
    Longitudinal Velocity & meters/second & Longitudinal instantaneous velocity of vehicle. \\
    Lateral Acceleration & meters/second$^2$ & Lateral instantaneous acceleration of vehicle.\\
    Longitudinal Acceleration & meters/second$^2$ & Longitudinal instantaneous acceleration of vehicle.\\
    Space Headway & meters & The distance between the front-center of a vehicle to the front-center of the preceding vehicle.\\
    Time Headway & seconds & The time to travel from the front-center of a vehicle (at the speed of the vehicle) to the front-center of the preceding vehicle.\\
    \bottomrule
    \end{tabular}
    \caption{Vehicle features used to constitute a DOP.}     \label{tab:DOP}
\end{table*}

\subsection{Analysis of Vehicle Lane Change}
Apart from the driving style, the other key factors of determining lane change are speed gain, headway tolerance, and safety. The main motivation of DLC is to gain faster speed and better driving conditions. Drivers are more likely to make lane-change decisions when lane change brings faster speed or larger space headway, or the headway to the preceding vehicle is no longer comfortable. Besides, safety is the prerequisite of lane-change decisions. Drivers will not change lanes if the traffic condition is perceived as unsafe for taking such an action. Next, we describe these three factors, \textbf{speed gain}, \textbf{tolerance}, and \textbf{safety}, in detail.

Denote $v_i, d_i, i \in \{E,\textit{P},  \textit{PL}, \textit{PR}, \textit{FL}, \textit{FR}, \textit{ASL}, \textit{ASR} \}$ as the velocity of the vehicle $i$ and the distance of the ego vehicle to the vehicle $i$, respectively. $v_i$ and $d_i$ will be set as $0$ if there is no such a vehicle $i$.
\begin{itemize}
    \item \textbf{Speed gain.} The speed gain obtained by the vehicle is the difference of the reachable speed in the current lane and the reachable speed in the target lane with respect to the current speed. The reachable speed in a lane is affected by the speed of the preceding vehicle in this lane, and the longitudinal distance to the preceding vehicle in this lane. Since our model can make decision of left lane change, right lane change, and lane keep, the speed gain factors can be captured by $(v_E-v_P), (v_{PL}-v_P), (v_{PR}-v_P)$, $(d_{PL}-d_P), (d_{PR}-d_P)$. 
    \item \textbf{Safety.} To ensure the safety of lane change, the driver must assess the collision risk with the following vehicle in the target lane. The collision risk will be low when the distance to the following vehicle is large enough for the ego vehicle to finish the lane-change actions. Meanwhile, the collision risk is related to the velocity difference between the ego vehicle and the following vehicle in the target lane, i.e., the collision risk is high when the speed of the ego vehicle is lower than the speed of the following vehicle in the target lane. The safety factors can be captured by $d_{FL}, d_{FR}, (v_E-v_{FL}), (v_E-v_{FR})$.
    \item \textbf{Tolerance.} The tolerance factor is used to measure the driving condition in the current lane, which is highly related to the safe time headway $t_h$ of the ego vehicle and the distance to the preceding vehicle. Specifically, drivers prefer to keep a distance longer than the safe distance on a highway to improve the driving comfort level. Thus, we use $(d_P-v_E\cdot t_h)$ as the variable of the tolerance factor.
\end{itemize}
In summary, as part of the input of the DSA-DLC decision-making model, these ten features $(v_E-v_P), (v_{PL}-v_P), (v_{PR}-v_P), (d_{PL}-d_P), (d_{PR}-d_P), d_{FL}, d_{FR}, (v_E-v_{FL}), (v_E-v_{FR}), (d_P-v_E\cdot t_h)$ are used to represent the traffic factors that affect the lane-change decision.

\subsection{Model Design}
In the aforementioned analysis, we have listed all variables that affect the lane-change decision, including driving style, speed gain, safety, and tolerance. The lane-change decision-making model can then be formulated as  
\begin{equation}
\begin{aligned}
    y =& f_\theta (v_E-v_P, v_{PL}-v_P, v_{PR}-v_P, d_{PL}-d_P,\\
    &  d_{PR}-d_P, d_{FL}, d_{FR}, v_E-v_{FL}, v_E-v_{FR}, \\ 
    & d_P-v_E\cdot t_h, D)
\end{aligned}
\end{equation}
where $y$ is a lane-change decision vector with three elements, corresponding to the probabilities for lane keep, left lane change, and right lane change. The lane-change decision-making is a multi-parametric and nonlinear problem that is difficult to solve through traditional mathematical methods. In recent years, deep learning shows great potential to solve this type of problem. Thus, we leverage deep learning models to solve this problem. With the trajectory dataset, we can extract the input variables and the decisions frame-by-frame, and then model it as a supervised learning problem. The learning objective is to find a model of $f_\theta(.)$ to minimize the long-term average loss defined by:
\begin{equation} \label{eq:loss}
    \frac{1}{N} \large{\sum_{i=1}^N} L(\hat{y}_i, f_{\theta}(.)) = \frac{(-1)}{N} \large{\sum_{i=1}^N} \sum_{c=1}^3 \hat{y}_{ic} \log(y_{ic}),
\end{equation}
where $N$ is the number of input-output pairs for training, $\hat{y}_i$ is the real decision vector obtained from the dataset, and we use the cross-entropy function as the loss function $L$, in which $\hat{y}_{ic}=1$ if the actual decision is $c$, and $\hat{y}_{ic}=0$ otherwise.

The DSA-DLC decision-making model aims to use the above variables as input to generate the lane-change decision. It consists of convolutional neural networks (CNNs) and fully connected (FC) neural networks. Currently, the CNN shows a great power on image classification in capturing the hidden feature of an image. The driving style of a vehicle is regarded as the hidden feature of the vehicle, and CNNs are used to extract the hidden driving style from the DOP. Since the impacts of the driving styles of the ego vehicle and the surrounding vehicles on the lane-change decision are different, we use two CNNs, one to deal with the DOP of the ego vehicle, and the other to deal with the DOPs of surrounding vehicles. Moreover, as the driving styles of surrounding vehicles have mutual influences on each other, we combine the DOPs of the surrounding vehicles into a 7-channel matrix and then use CNNs to extract the hidden patterns. The driving styles of the ego vehicle and the surrounding vehicles are represented as vectors after being processed by CNNs.

The variables of speed gain, safety, and tolerance are represented as a 10-dimension vector. The two driving style vectors output from CNNs are concatenated with the 10-dimension vector and then input to the FC networks. Finally, the model can output a 3-class result, i.e., lane keep, left lane change, or right lane change. We use Eq.~\ref{eq:loss} to calculate the loss value and optimize all of the neural networks.

\section{Evaluation and Analysis} \label{sec:experiments}

To evaluate the DSA-DLC decision-making model, we choose the highD dataset~\cite{highDdataset} to train the model. Specifically, we evaluate the model from two perspectives: (1) the accuracy with respect to human drivers' decisions, and (2) the lane-change impact on the following vehicle in the target lane. In the rest of this section, we first introduce the dataset used to train and test our model. Then we present the data pre-processing procedure. Next, we elaborate on the parameters and architectures of the model, and illustrate the accuracy. Finally, we discuss and analyze the lane-change impact of using our model.

\subsection{Dataset} 

The highD dataset is a vehicle trajectory dataset recorded by drones at six different Germany highways. It records $110,500$ vehicles' data, including trajectory, type, and size. It covers around $2520$-meter highway for $16.5$ hours in total. Furthermore, the highD dataset is recorded in different times of the day, covering various density of traffic flows. The typical speeds for trucks and cars at the recording sites are $80$ km/h and $120$ km/h, respectively. Compared to the widely-used next generation simulation dataset (NGSIM) datatset~\cite{NGSIM}, highD contains nearly eleven times more vehicles, and eight times longer travelling distance. The speeds of cars and trucks in highD are closer to the real highway traffic flow and are much higher than those in NGSIM. The highD dataset includes more than $11,000$ lane changes, which is twice the number of NGSIM. In addition, the highD dataset has higher quality than NGSIM. Specifically, the NGSIM dataset includes many errors and the accuracy of trajectories affects the lane-change model parameters~\cite{montanino2015trajectory}. Therefore, we choose the highD dataset for our model's performance evaluation.  

\begin{figure}
    \centering
    \includegraphics[width=10cm]{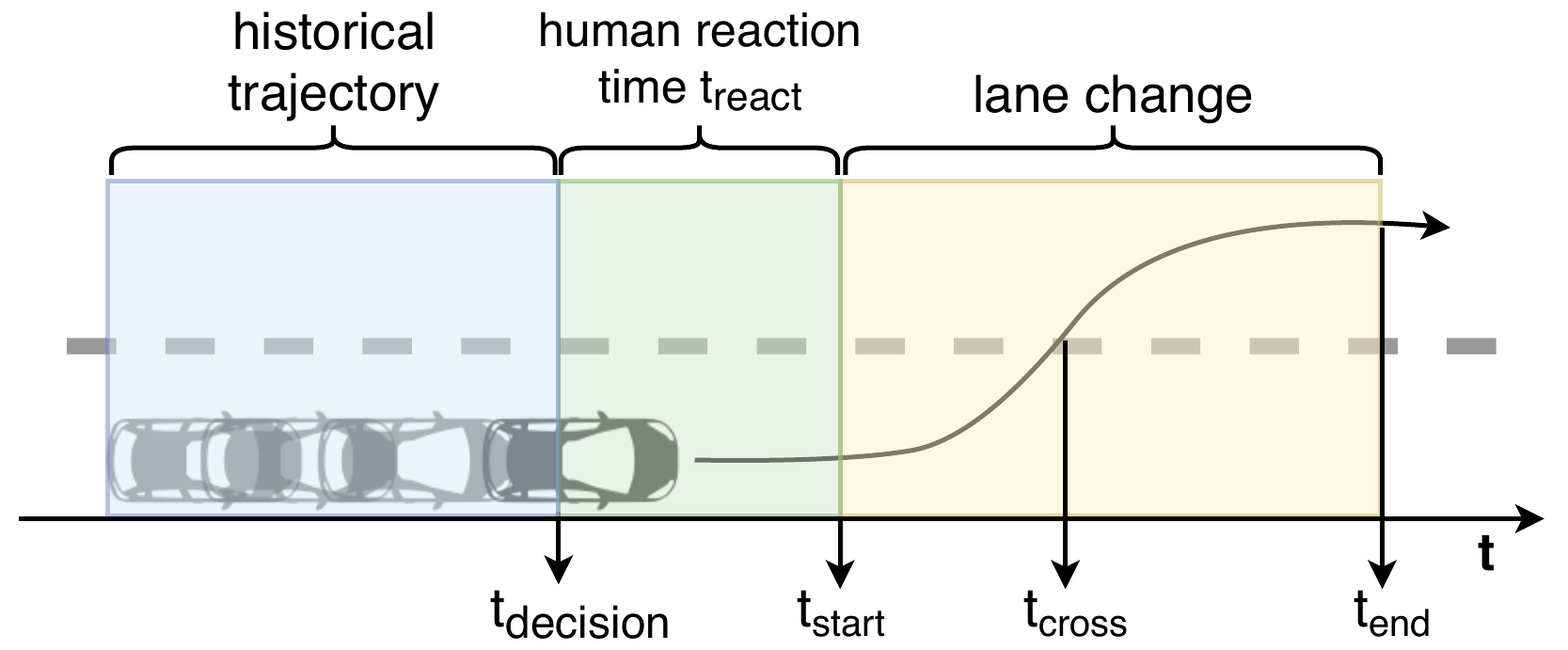}
    \caption{Lane-change process.}
    \label{fig:data_extraction}
\end{figure}

\begin{table*}[t]
    \centering
    \setlength{\tabcolsep}{0.85em} 
{\renewcommand{\arraystretch}{1.1}
    \begin{tabular}{|p{3.8cm}<{\centering}|c|c|c|c|c|}
    \hline
    Model & \diagbox{Real}{Predict} & Keep & Left & Right & Overall Accuracy\\
    \hline
    \multirow{3}*{\shortstack{With DS of the ego vehicle\\and surrounding vehicles}} & Keep & \textbf{98.91\%} & 0.52\% & 0.57\% & \multirow{3}*{\textbf{98.66\%}}\\
    \cline{2-5}
      & Left & 3.27\% & \textbf{96.73\%} & 0.0\% & \\
    \cline{2-5}
       & Right & 3.21\% & 0.0\% & \textbf{96.79\%} & \\
    \hline
    \multirow{3}*{\shortstack{Without DS of \\the ego vehicle}} & Keep & 88.20\% & 6.24\% & 5.56\% & \multirow{3}*{87.54\%}\\
    \cline{2-5}
      & Left & 12.15\% & 85.98\% & 1.87\% & \\
    \cline{2-5}
       & Right & 16.79\% & 3.57\% & 79.64\% & \\
    \hline
    \multirow{3}*{\shortstack{Without DS of \\ surrounding vehicles}} & Keep & 94.80\% & 0.44\% & 4.75\% & \multirow{3}*{95.12\%}\\
    \cline{2-5}
      & Left & 4.67\% & 95.33\% & 0.0\% & \\
    \cline{2-5}
       & Right & 0.71\% & 0.0\% & \textbf{99.29\%} & \\
    \hline
    \multirow{3}*{\shortstack{Without any DS}} & Keep & 84.10\% & 7.09\% & 8.81\% & \multirow{3}*{83.61\%}\\
    \cline{2-5}
      & Left & 15.42\% & 81.78\% & 2.80\% & \\
    \cline{2-5}
       & Right & 18.57\% & 3.22\% & 78.21\% & \\
    \hline
    \end{tabular}}
    \caption{Comparison among different models.}
    \label{tab:our_model}
\end{table*}

\subsection{Data Pre-processing}
We first need to extract all DLC cases from the trajectories. Since the vehicles from the on-ramp need to make mandatory lane change to the main lanes, we eliminate all these lane changes from the dataset. After that, the number of DLC cases in the dataset is $9249$, including $4099$ left lane changes and $5150$ right lane changes. There are four critical moments for a lane change as shown in Figure~\ref{fig:data_extraction}: 
\begin{itemize}
    \item $t_{decision}$: the time when the vehicle makes the lane-change decision,
    \item $t_{start}$: the time when the vehicle starts taking actions for changing lane,
    \item $t_{cross}$: the time when the vehicle crosses the line,
    \item $t_{end}$:  the time when the lane change is finished and the vehicle keeps traveling in the target lane.
\end{itemize}
Since the decision-making moment is hard to recognize, we infer 
$t_{decision}$ with $t_{start}-t_{react}$, where $t_{start}$ is obtained according to the automatic labeling method in~\cite{zhang2020novel} and $t_{react}$ is human's reaction time. According to~\cite{reaction_time}, $t_{react}$ varies from $0.7$ to $1.5$ seconds, so we select the average value, i.e., $1$ second, in our experiments. The lane-change decision is made according to the observations of $7$ surrounding vehicles, the driving style of the ego vehicle, and the aforementioned traffic factors. Thus, for each lane-change vehicle, we extract the $2$-second historical trajectory of its own and its surrounding vehicles, i.e., the data from the $(t_{decision}-2)$ second to the $t_{decision}$ second, to constitute the lane-change data, since DOPs are calculated using $2$-second trajectory data. 

Besides the lane-change data, we also need to retrieve the lane-keep data for model training and testing. To avoid the noise in the lane-keep trajectory after a lane change, we select the vehicles that stay in the same lane for longer than $12$ seconds as lane-keep vehicles. Following the lane-change data, we segment the trajectory of lane-keep vehicles and their surrounding vehicles into $2$-second time windows to constitute the lane-keep data. Each lane-change/keep vehicle's $2$-second trajectory with its surrounding vehicles' trajectories is defined as a lane-change/keep case.

Given the above labeled lane-change/keep data, we construct DOPs and calculate the value of the aforementioned traffic factors for model input. We randomly select $90\%$ data as the training set and the rest $10\%$ data as the test set. As the lane-keep data dominates the training data, we duplicate $16$ times lane-change data to balance the training data. Note that the lane-change data in the test set is not duplicated to ensure the reliability of the performance evaluation.

\subsection{Network Architecture and Training Parameters}
The CNN for capturing the surrounding vehicles' driving styles is called CNN1 and the other CNN for capturing the ego vehicle's driving style is called CNN2. CNN1 contains $2$ convolutional layers where the first one has $16$ kernels with size $4$ and the second one has $32$ kernels with size $5$. CNN2 also includes $2$ convolutional layers where the first one has $16$ kernels with size $4$ and the second one has $8$ kernels with size $5$. There are $4$ FC layers, including $50$ neurons for the first layer, $128$ for the second one, $32$ for the third one, and $16$ for the last one, respectively. We choose the Adam optimizer with a learning rate of $0.001$ and a batch size of $16$ to train the model for $50$ epochs. 

\subsection{Numerical Results}
The overall accuracy of the model is $98.66\%$ and the details are illustrated in Table~\ref{tab:our_model}. The accuracy of lane keep, left lane change, and right lane change are $98.91\%$, $96.73\%$, and $96.79\%$, respectively. Only $1.09\%$ of lane-keep cases are predicted as left/right lane change. The $3.27\%$ left lane changes and $3.21\%$ right lane changes are predicted to be lane keep. Through analysis of the wrongly predicted cases, we find that our model tends to be a relatively ``conservative driver", which prefers lane keep than lane change. 

To illustrate the impacts of driving styles (DS), we conduct the following three experiments for comparison.
\begin{itemize}
    \item \textbf{Eliminate the driving style of the ego vehicle.} The DOP of the ego vehicle is not input to the model and the corresponding neural networks for processing this DOP are disabled.
    \item \textbf{Eliminate the driving style of surrounding vehicles.} The DOPs of surrounding vehicles are not input to the model and the corresponding neural networks for processing these DOPs are disabled.
    \item \textbf{Eliminate the driving style of the ego vehicle and surrounding vehicles.} Both the DOPs of the ego vehicle and surrounding vehicles are not input to the model and the corresponding neural networks for processing these DOPs are disabled.
\end{itemize}

The comparison results are shown in Table~\ref{tab:our_model}. The accuracy only achieves $83.61\%$ when there is no driving style information at all, which is the lowest one. In contrast, the accuracy is higher when we add a part of driving style information. Specifically, as illustrated in the experimental results, the driving style of the ego vehicle has more impact on the final decision. The accuracy without considering the driving style of the ego vehicle is $87.54\%$ which is lower than the accuracy without adding the driving style of surrounding vehicles. When the driving style of the ego vehicle is input to the model, the right lane-change prediction accuracy achieves $99.26\%$. However, the overall accuracy, especially for predicting lane keep, performs poorer than our original model that considers all driving styles. Therefore, it is imperative to consider the driving style to determine the lane-change decision.

\subsection{Lane-change Impact Analysis}
Besides the accuracy, a more important metric for evaluating a lane-change decision-making model is the impact of lane change on traffic. We evaluate our DSA-DLC decision-making model by analyzing the impact of lane change on the following vehicle in the target lane from two perspectives: (1) speed and (2) safety. 

From Table~\ref{tab:our_model}, $6.48\%$ lane-change cases are predicted as lane keep (LK) while only $1.09\%$ of lane-keep cases are predicted as lane change (LC). Thus, we focus on the analysis of the former cases with a higher wrongly predicted ratio. To better visualize the performance, we use a confusion matrix shown in Table~\ref{tab:confusion_matrix} to represent all cases. The rest of this section demonstrates that the lane change decisions made by human drivers have more negative impacts on traffic than our model by comparing the TN cases and FP cases.

\begin{table}
\begin{center}
\def\mathbi#1{\textbf{\em #1}}
\setlength{\tabcolsep}{0.85em} 
{\renewcommand{\arraystretch}{}
\begin{tabular}{@{}lll@{}}
\toprule
 & \textbf{Actual LK}  & \textbf{Actual LC}  \\ \midrule
\textbf{Predict LK} & True Positive (TP)  & False Positive (FP) \\ \midrule
\textbf{Predict LC} & False Negative (FN) & True Negative (TN)  \\ \bottomrule
\end{tabular}}
\caption{Confusion Matrix.}
\label{tab:confusion_matrix}
\end{center}
\end{table}

\begin{table}
\begin{center}
\def\mathbi#1{\textbf{\em #1}}
\setlength{\tabcolsep}{0.8em} 
{\renewcommand{\arraystretch}{1.1}
\begin{tabular}{@{}ccccc@{}}
\toprule
Speed Change Rate  & 0 & -1\% & -2\% & -3\%  \\  \midrule
Percentage in TN cases  & 43.16\% & 34.47\% & 24.47\% & 17.37\% \\ \midrule
Percentage in FP cases  & 43.86\% & 35.96\% & 28.07\% & 21.93\% \\ \bottomrule
\end{tabular}}
\caption{Comparison of speed change rate of the following vehicle in the target lane between TN cases and FP cases.}
    \label{tab:speed_change_rate}
\end{center}
\end{table}

\textbf{Speed Change Rate.} According to~\cite{YANG2019317}, the speed change rate of the following vehicle in the target lane can well measure the lane-change impact, which is represented as 
\begin{equation}
        speed\ change\ rate = \frac{v_{end} - v_{start}}{v_{start}} \times 100\%
\end{equation}
where $v_{start}$ and $v_{end}$ are the speeds of the following vehicle in the target lane when the lane-change vehicle starts and completes changing lane, respectively. The negative value of this rate defines the fluctuation of the speed and deceleration, which indicates a negative impact of this lane change. The higher absolute value of this rate indicates a larger fluctuation, which decreases passengers' comfort.

We calculate the percentage of the following vehicle in the target lane whose speed change rate is negative and less than a given value. The results are illustrated in Table~\ref{tab:speed_change_rate}. From the table, the percentage of speed change rate less than $-1\%$ for these two types of cases are almost the same. However, $21.93\%$ following vehicles in the target lane are forced to decelerate more, i.e., the speed change rate is less than $-3\%$, when our model predicts LK but actual LC. In contrast, there are only $17.37\%$ following vehicles in the target lane whose speed change rate is less than $-3\%$ when our model also predicts to change lanes. Therefore, the lane change decision made by our model has less negative impacts on traffic.

\begin{table}
\begin{center}
\setlength{\tabcolsep}{0.8em}
\begin{tabular}{@{}cccc@{}}
\toprule
Safety Impact& Positive & Negative & None \\ \midrule
Percentage in TN cases & 23.60\% & 31.37\% & 45.03\%  \\ \midrule
Percentage in FP cases & 21.21\% & 41.41\% & 37.38\%  \\ \bottomrule
\end{tabular}
\caption{Comparison of the safety impacts on the following vehicle in the target lane between TN cases and FP cases.}
\label{tab:ttc_change}    
\end{center}
\end{table}

\textbf{Safety.} We use the time-to-collision (TTC) to measure safety. The TTC is defined as
\begin{equation}
    \mathrm{TTC} = \frac{x_l - x_f}{v_f - v_l}
\end{equation}
where $x_l$ and $x_f$ are the coordinates of the leading vehicle and the following vehicle, respectively, and $v_l$ and $v_f$ are the velocities of the leading vehicle and the following vehicle, respectively~\cite{kruber2019highway}. The value of TTC in the highD dataset can be positive, negative, or zero. Positive TTC indicates the collision will occur after TTC seconds if the leading vehicle and the following vehicle keep traveling with the current speed. Negative TTC indicates no collision, and zero TTC indicates no preceding vehicle. The TTC of the following vehicle in the target lane will be influenced by lane change. We then analyze the safety impact, i.e., positive, negative, and no impact, through comparing the TTC of the following vehicle in the target lane before the lane change starts (denoted as $\mathrm{TTC}_{start}$) and its TTC after the lane change ends (denoted as $\mathrm{TTC}_{end}$) in the following:

\begin{itemize}
    \item $\mathrm{TTC}_{start} > 0$ and $\mathrm{TTC}_{end} < 0$: the lane change brings \textbf{positive} impacts.
    \item $\mathrm{TTC}_{start} \leq 0$ and $\mathrm{TTC}_{end} > 0$: the lane change brings \textbf{negative} impacts.
    \item $\mathrm{TTC}_{start} \leq 0$ and $\mathrm{TTC}_{end} < 0$: the lane change brings \textbf{no} impact.
    \item $\mathrm{TTC}_{start} > 0$ and $\mathrm{TTC}_{end} > 0$: in this case, we use $\mathrm{TTC}_{end} - \mathrm{TTC}_{start}$ to measure the impact. Specifically, the positive value indicates a \textbf{positive} impact, and the negative value indicates a \textbf{negative} impact. 
\end{itemize}

As shown in Table~\ref{tab:ttc_change}, $41.41\%$ lane changes bring negative impacts to the following vehicle in the target lane in FP cases while the negative impact percentage is $31.37\%$ in TN cases. That is, our model has learned the safety pattern between the ego vehicle and the following vehicle in the target lane, and then made lane-keep decisions in FP cases to mitigate the negative impact on traffic safety.


\section{Conclusion and Future Work} \label{sec:conclusion}
In this paper, we proposed and implemented a DSA-DLC decision-making model to ``teach'' the machine to make lane-change/keep decisions like a human driver, which is the first model considering both the driving styles of surrounding vehicles and the ego vehicle. Specifically, the hidden risks from surrounding vehicles are mitigated by considering their driving styles. The driving style of the ego vehicle helps to imitate human drivers' decision-making maneuvers. Combining the above information with the contextual traffic information, our model can make a similar decision as human drivers under the same circumstance, which makes our model ``human-like", an important and desirable feature in autonomous driving to facilitate a smooth transition from the current transportation systems to the future transportation systems where the autonomous driving is the mainstream. Data-driven experiments show that our model can achieve a high accuracy of $98.66\%$ with respect to human drivers' decisions. In addition, the model can simultaneously improve the safety and efficiency of traffic.


In future work, we plan to apply our model in traffic flow simulators to simulate a human-like traffic flow and study the impact of autonomous vehicles on human-driven vehicles. It is acknowledged that the mixed traffic flow containing both human-driven vehicles and autonomous vehicles will be common in the coming decades. Therefore, to gain public acceptance and increase the market penetration of autonomous vehicles, it is crucial to study the impact of autonomous vehicles on the human-driven traffic flow and, more importantly, to leverage such impacts to positively influence the human-driven vehicles so as to improve traffic efficiency and safety. To this end, simulating human-like traffic flow is an important step to achieve the above goal.

\bibliographystyle{unsrt}  


\end{document}